\pdfoutput=1

\documentclass[11pt]{article}

\usepackage[preprint]{acl}

\usepackage{times}
\usepackage{latexsym}

\usepackage[T1]{fontenc}

\usepackage[utf8]{inputenc}

\usepackage{microtype}

\usepackage{inconsolata}

\usepackage{graphicx}

\usepackage[table]{xcolor}
\usepackage{multirow}

\usepackage{makecell}
\usepackage{ragged2e}
\usepackage[dvipsnames]{xcolor}

\usepackage{linguex} 
\usepackage{nameref}
\usepackage{arydshln}
\usepackage{ulem}
\usepackage{tikz}
\usepackage[most]{tcolorbox}
\usepackage{xcolor} 

\title{Shortcomings of LLMs for Low-Resource Translation: Retrieval and Understanding are Both the Problem}
\author{Sara Court \\
The Ohio State University \\
  {\tt court.22@osu.edu} \\\And
  Micha Elsner \\
  The Ohio State University \\
  {\tt elsner.14@osu.edu}}

\begin{document}
\maketitle
\begin{abstract}
This work investigates the in-context learning abilities of pretrained large language models (LLMs) when instructed to translate text from a low-resource language into a high-resource language as part of an automated machine translation pipeline. We conduct a set of experiments translating Southern Quechua to Spanish and examine the informativity of various types of context retrieved from a constrained database of digitized pedagogical materials (dictionaries and grammar lessons) and parallel corpora. Using both automatic and human evaluation of model output, we conduct ablation studies that manipulate (1) context type (morpheme translations, grammar descriptions, and corpus examples), (2) retrieval methods (automated vs. manual), and (3) model type. Our results suggest that even relatively small LLMs are capable of utilizing prompt context for zero-shot low-resource translation when provided a minimally sufficient amount of relevant linguistic information. However, the variable effects of context type, retrieval method, model type, and language-specific factors highlight the limitations of using even the best LLMs as  translation systems for the majority of the world's 7,000+ languages and their speakers.
\end{abstract}

\section{Introduction}
Despite great progress in the quality of today's state of the art machine translation (MT) systems, constraints on the amount and kinds of data available in the majority of the world's 7,000+ languages have led to yet another disparity in access and support for speakers of these languages: low-resource MT continues to be a major challenge \cite{hendy2023good, nicholas2023lost, robinson-etal-2023-chatgpt, stap-araabi-2023-chatgpt}. Although many languages lack the kinds of large, standardized corpora necessary for traditional MT methods, recent work suggests it may be possible to leverage a smaller amount of existing resources, for example pedagogical materials used for language instruction, to develop MT systems with Large Language Models (LLMs), albeit with varying results \cite{elsner-needle-2023, tanzer2024mtob,zhang2024teaching}.
These materials are often the result of community-driven or government-led initiatives to support language revitalization, reclamation, and mother-tongue education \cite{schreiner-etal-2020-multidirectional,liu-etal-2022-notalways, riestenberg-freemond-2024-prioritizing}. Such discrepancies in the needs and priorities of
academic, commercial, and community-led 
efforts to develop digital resources and language technologies is what \citet{gessler-2022-nlpgap} terms the ``NLP Gap''.


In this study, we investigate one way to lessen the NLP Gap,
comparing LLMs' in-context learning abilities when translating from a low-resource language (a Peruvian variety of Southern Quechua) to a high-resource language (Spanish) using information retrieved from a database of pedagogical materials. We replicate results of earlier studies on a new language pair by comparing the effects of morpheme translations, sentences from a parallel corpus, and passages from a grammar instruction document on translation quality. We then conduct a more focused analysis by annotating translation outputs by hand using a modified MQM error typology \cite{burchardt-2013-mqm}. Finally, we conduct an ablation study on the effects of automated retrieval by manually constructing prompts using the same set of materials.

Our results suggest that while, unsurprisingly, translation quality improves with model size, such improvements seem to primarily be the result of previous exposure to the low-resource language during model pretraining, rather than an improved ability for the model to utilize prompt context, as evidenced by high scores in response to baseline (zero-shot) translation prompts. However, we also find evidence that in-context learning abilities may be inconsistent across different models of similar size. As found in previous studies, prompts containing morpheme and word-level translations reliably improve model outputs, but information from the grammar and corpus have a null or even negative effect on results. 
Human evaluation on a selection of outputs from two models -- GPT-3.5 Turbo and GPT-4o -- 
align with the quantitative measures we obtain using BLEURT \cite{sellam2020bleurt} as an automatic metric. Quantitative results also show an effect of automated retrieval on translation quality that is most evident in prompts containing morpheme translations and for models with lower baseline scores. Finally, we highlight a number of ethical concerns and limitations that arise from the proposed methods that are supported by our findings, and discuss the potential risks and challenges LLM-based methods for low-resource MT face moving forward.

\section{LLMs for Machine Translation}


Modern LLMs are now capable of translating many high-resource languages, but lack sufficient coverage of even modestly resourced languages to achieve comparable results without additional support \cite{kocmi-etal-2023-findings}.
Retrieval-augmented generation \cite{rubin-etal-2022-retrieval}
may provide such support in the form of parallel sentences \cite{agrawal2022incontext}, dictionary definitions \cite{ghazvininejad2023dictionarybased,lu2023chainofdictionary} or other linguistic meta-knowledge such as a grammatical description.
Retrieval-augmented methods offer exciting possibilities for low-resource translation, since the LLM might (in principle) be able to ``teach itself'' the language from learner-oriented resources produced by community members or language specialists.

Studies to date \cite{elsner-needle-2023, reid2024gemini,zhang2024teaching} experiment with four dimensions of variability: source language, LLM, type(s) of information retrieved, and retrieval method. 
Since the source languages in these studies have relatively little presence in public corpora or on the web, differing results across LLMs can tentatively be attributed to differences in their in-context learning and instruction-following abilities.

All studies find that word-level translations are helpful additions to prompts. \citet{zhang2024teaching} and \citet{tanzer2024mtob} also add sentence pairs from a parallel corpus, while \citet{elsner-needle-2023} add usage examples from a dictionary. Each improves results, although to a lesser degree. \citet{elsner-needle-2023} and \citet{zhang2024teaching} experiment with small fixed ``grammar lesson'' passages to provide explicit syntactic instruction, but find these ineffective. \citet{tanzer2024mtob} uses passages retrieved from a grammar book, also with relatively disappointing results. \citet{reid2024gemini} use the entire grammar book and a very long-context model to obtain better translations, but without exploring the role explicit grammar instruction actually plays in doing so.

\citet{zhang2024teaching} find that sentences from the corpus retrieved using BM25 embeddings \cite{robertson2009bm25} work better than random ones. \citet{tanzer2024mtob}, however, report that retrieval with longest common substring (LCS) matching outperforms embedding-based retrieval. Overall, the question of how to best retrieve relevant passages containing grammar material or sentences in a low-resource language is still open. This also complicates the interpretation of the mostly-negative results found for grammar passages. It is not clear whether these stem from poor retrieval, from the LLM's inability to process the retrieved content, or both. Moreover, although \citet{reid2024gemini} conducts human evaluation of the results for quality, to the best of our knowledge no study to date systematically investigates specific grammatical errors in the output. 

Finally, each of these studies finds a significant decrease in LLMs' abilities to translate from a high-resource language into a low-resource language relative to experiments in the opposite direction. This is in line with \citet{mccoy2023embers}, who find that while the accuracy of an LLM's output highly depends on the probability of both the input and the output text, output probability has a greater influence on model performance.
We therefore focus this study on a single translation direction, instructing LLMs to output translations from a low-resource language into the language with which they are likely to have had more exposure during training, i.e, from Southern Quechua into Spanish, and leave the reverse direction for future work.

\section{Quechuan Languages}
Quechua is a family of languages Indigenous to the Andes in South America. This study focuses on varieties of Southern Quechua (S. Quechua, also known as \textit{urin quechua} or \textit{quechua sureño}) spoken in parts of Peru.\footnote{Unless noted otherwise, we use \textit{Quechua} in this study to refer Southern Quechua and related varieties.}
While previous studies investigated language-LLM pairs for which the baseline LLM lacked any pretrained knowledge, we find that newer LLMs can translate some S. Quechua sentences in a zero-shot setting. We expect this to be typical of many low-resource languages which, while often endangered, still may have some presence on the web.

Quechuan languages have by far the largest representation of all Indigenous Latin American languages in NLP research \cite{tonja2024indig-nlp} and are often included in ACL-affiliated workshops, datasets, and shared tasks \cite{cotterell2020conllsigmorphon, ebrahimi-etal-2022-americasnli, ebrahimi-etal-2023-findings}.
S. Quechua has a robust language toolkit \cite{rios2015quechuatoolkit}, including the morphological parser we use in our pipeline. It has also been the subject of numerous studies on MT for both text and speech, developed in conjunction with monolingual and parallel corpora \cite{rios2015quechuatoolkit, cardenas2018siminchik, ortega2020neural, zevallos-etal-2022-qubert}. Nonetheless, such tools continue to face challenges, and Quechuan languages continue to lack the resources necessary to develop most of today's state of the art models. 

Since Quechua is primarily spoken in South America, the majority of available digital resources, including all materials used in this study, use Spanish as the language of translation, explanation, and/or instruction. We therefore also use Spanish, rather than English or any other high-resource language, as the language of translation and prompting when testing our system.

\subsection{Language-Specific Factors}
\label{sec:morphology}

While the proposed methods are general enough to be applied to any language pair, model outputs may reflect certain language-specific characteristics of the source and target languages, respectively. In this section, we provide a brief description of selected language-specific factors in S. Quechua as they relate to their translated Spanish counterparts. For a discussion of their potential effects on our results, please see Section \ref{sec:lang-effects}.

\subsubsection{Morphological Segmentation}
S. Quechua is primarily agglutinating, i.e., much of the morphology may be described in terms of isomorphic form-meaning relationships, morphemes generally maintain a consistent form regardless of their phonological environment, and morpheme boundaries tend to be transparent. In contrast, morpheme segmentation in Spanish may be rendered opaque due to its fusional morphology and widespread use of conditioned allomorphy.

While LLMs are trained to process text via token-based rather than morpheme-based segmentation, it is possible that a lack of direct correspondence between the expression of morphosyntactic categories in S. Quechua and Spanish may affect a model's ability to leverage the information we provide as prompt context in our experiments. Correspondences in form and meaning across parallel usage examples may be particularly obscured, limiting the use of corpora designed for traditional MT methods. It may be possible to mitigate such issues with more advanced retrieval or prompting techniques, for example by explicitly instructing an LLM to conduct morphological analysis as part of the translation process, but we leave this for future work.

\subsubsection{Syncretism and Polysemy}
Although the language is primarily agglutinating, a number of morphemes in S. Quechua are syncretic, such that a given form may be used to express more than one grammatical category. For example, the \textsc{1sg.poss} marker, \textit{-y}, shares the same form as both the \textsc{2sg.imp} marker and the infinitival marker, as illustrated in the following examples:

\exg. ñaña-y\\
sister-\textsc{1sg.poss}\\
`my sister'\\
\label{data:y}

\exg. Mikhu-y!\\
eat-\textsc{2sg.imp}\\
`Eat!'\\

\exg. mikhu-y-ta muna-ni\\
eat-\textsc{inf}-\textsc{acc} want-\textsc{1sg}\\
`I want to eat'\\

Similarly, words in S. Quechua may be polysemous, with the potential to express more than one meaning depending on their use in context. For example, the S. Quechua word \textit{miski} (\textit{misk'i}) may be translated as either \textit{dulce} `sweet' or \textit{rico/a} `delicious', and both \textit{dulce} and \textit{rico/a} are themselves polysemous in Spanish. \textit{Dulce} may be used as an adjective, i.e., `sweet', or a noun, i.e.,`candy', and \textit{rico/a} may describe either richness in flavor, i.e., `delicious', or in monetary wealth, i.e., `rich'.

The exact forms displaying syncretism or polysemy must be identified on a language-specific basis, but the ambiguity they present poses a clear problem for our proposed methods in general, with potential effects on both retrieval and generation. We discuss this issue further in Section \ref{sec:lang-effects}.

\subsubsection{Variation}
Both S. Quechua and Spanish are characterized by extensive regional and dialectal variation. In S. Quechua,  this includes differences in orthographic and/or phonological conventions as well as the specific lexical items and expression of morphosyntactic content. For example, the S. Quechua word for `dog' may be rendered orthographically as \textit{alqo}, \textit{allqo}, \textit{allku}, \textit{allqu}, or \textit{ashko}, and the additive suffix may be expressed as either \textit{-pas} or \textit{-pis}, depending on the community. 
Variation in the attested usage of specific lexical items and morphemes across communities is also common in S. Quechua. For example, the evidential marker \textit{-mi} /\textit{-m} is frequently attested in the Peruvian variety of S. Quechua used in this study, but essentially absent in many Bolivian varieties. 

Variation across Spanish-speaking communities may also affect models' abilities to produce translations that are both accurate and appropriate. The Andean Spanish reference translations used in this study do not appear to affect the results of our automatic evaluation. However, were the proposed methods to be applied in a realistic setting, it would be especially important to assess the degree of alignment between any prescriptive linguistic standards that have been implicitly acquired by the LLM and the usage conventions of the language community or communities of interest.



\section{Methods}
\subsection{Data}
We conduct experiments on a collection of 50 pairs of S. Quechua - Spanish sentences sourced from one of the author's personal notes. These were selected to highlight a range of specific grammatical phenomena at multiple levels of difficulty--- they include simple clauses and tenses (Example \ref{data1}), as well as more advanced constructions such as those involving past participles (Example \ref{data2}) and simultaneous events (Example \ref{data3}.

\exg. qam allin-ta tusu-nki\\
you good-\textsc{acc} dance-\textsc{2sg}\\
tu bailas bien \\
`you dance well'\label{data1}
\newline
\newline

\exg. awa-sqa-y wali-qa sumaq-mi\\
knit-\textsc{pp}-\textsc{1sg} skirt-\textsc{top} great-\textsc{assertive}\\
la falda que tejí es linda \\
`the skirt that I knit is pretty'\label{data2}

\exg. qam-qa taki-ta uyari-spa wasi-yki-ta picha-chka-nki\\
you-\textsc{top} song-\textsc{acc} listen-\textsc{subr} house-\textsc{2.poss}-\textsc{acc} clean-\textsc{prog}-\textsc{2sg}\\
tú estás limpiando tu casa escuchando música \\
`you're cleaning your house listening to music'\label{data3}

The first author, a foreign-language student of S. Quechua, received permission from her instructor to use notes from their lessons for the study. All sentence pairs were inspected by the instructor, a native bilingual speaker of both S. Quechua and Peruvian Spanish, to eliminate any errors and confirm the accuracy of all reference translations.

\subsection{Prompt Construction}
As a baseline, each sentence is inserted into a prompt template that instructs the model in Spanish to translate the S. Quechua sentence into Spanish and respond only with the translation (Figure \ref{fig:prompt}). We automate a process for building on this template and compare the effects of adding information from three different sources to the prompt context. 
\begin{figure}
        \small
        \begin{tcolorbox}[colback=gray!10, boxrule=0.1mm, width=\columnwidth]
        [TAREA] Traduce la siguiente frase del quechua al español. Responde sólo con la traducción: \\
        quechua: kay wasiqa turiypam \\
        español:
        \end{tcolorbox}
        \caption{Example \textsc{baseline} prompt. English: \textit{[TASK] Translate the following sentence from Quechua to Spanish. Respond only with the translation: Quechua: kay wasiqa turiypam; Spanish:}}
        \label{fig:prompt}
\end{figure}

\subsubsection{Morpheme Translations (\textsc{morph})}

We use a morphological parser \cite{rios2015quechuatoolkit} to segment each word of the source segment into morphemes, each with gloss symbols and a Spanish translation.\footnote{We set aside valid concerns regarding the theoretical status of the \textit{morpheme} for this study and define a morph(eme) loosely as a recognizable form-meaning pair that recurs in a language.} Some morphemes have multiple candidate meanings, all of which are retrieved. As an example, the word \textit{rantikuq} is segmented as \textit{ranti-ku-q} and glossed as ``comprar.DB.VRoot-DB.VDeriv.+RflxInt+Ag.NS.''

While numerous orthographic standards have been developed and promoted across Quechuan-speaking communities in South America, considerable variation in orthographic conventions may be found even within a particular community or variety \cite{rios2014morphological}. We discuss the implications of this for our results in Section \ref{sub:manual}.

We supplement the output 
from the parser using a Quechua-Spanish bilingual dictionary \cite{amlq2005-diccionario}. 
We retrieve any dictionary entry whose headword exactly matches a morpheme in our segmentation.
By default, we include all senses and any usage examples or contextual information in the dictionary entry as part of the prompt. We then concatenate the output of the parser with the retrieved dictionary entries and include this \textsc{morph} information as prompt context preceding the source sentence and baseline translation prompt.

\subsubsection{Grammar Descriptions (\textsc{grammar})}
We also experiment with the inclusion of grammar lessons found in student-facing pedagogical materials, retrieving grammatical explanations relevant to each source sentence from a PDF document developed for students and teachers of S. Quechua \cite{pinto2005didactica}.  
The document is organized into short sections (1-3 sentences, plus paradigm tables or usage examples) that describe the particular grammatical concept associated with an affix in Quechua.
For each source sentence, we retrieve sections associated with any affix listed in the document that is an exact match of a 
morpheme
and include this in prompts using contextual information from the grammar. This improves on the methods described in \citet{tanzer2024mtob}, who use LCS-based retrieval over an entire textbook, and \citeauthor{elsner-needle-2023} \citeyear{elsner-needle-2023} and \citeauthor{zhang2024teaching} \citeyear{zhang2024teaching}, whose grammatical description remains consistent across prompts regardless of the source text being translated.

\subsubsection{Parallel Usage Examples (\textsc{corpus})}
Finally, we experiment with sentence-level examples from a S. Quechua-Spanish parallel corpus designed for traditional NLP tasks. We combine data made available via the AmericasNLP 2021 Shared Task on Open Machine Translation and the 2023 IWSLT shared task on low-resource SLT \cite{tiedemann-2012-parallel, agic-vulic-2019-jw300, ortega2020neural, americasnlp2021, iwslt-2023-findings}. For each source sentence, we retrieve the three best matches from the corpus using a LCS search against the full source sentence.
\subsubsection{Combined Prompt Types}
Combinations of information from all three sources yields 8 total conditions, including the baseline. An example prompt from each information source is given in Appendix \ref{app:exampleprompts}.
 
\subsubsection{Manually Revised Prompts}
\label{sub:manual}
To compute a soft upper bound on the improvements possible with better retrieval, we conduct an additional set of experiments using manually revised prompts. We first examine the content retrieved from the morphological parser, dictionary, and grammar document and remove all instances of ambiguity and irrelevant or misleading information from the prompt context.\footnote{We do not experiment with retrieval methods for corpus examples, which were retrieved using LCS match in both conditions. Improving on LCS-based retrieval remains an open question in low-resource LLM-MT, and we leave this for future work.}

For example, many S. Quechua speakers use the term \textit{runasimi} (lit: `people mouth', `the people's language'), as an endonym for the language. The parser, however, returns only the literal decomposition (\textit{runa} `ser humanos'/`people' and \textit{simi} `boca'/`mouth'), and the dictionary does not list \textit{runasimi} as a headword but rather as one of eight different senses of \textit{simi}.
We thus remove all such irrelevant examples and translations from the prompt and retain only the content indicating a translation of \textit{runasimi} in the linguistic sense.

We also manually retrieve content from the dictionary and grammar documents that were overlooked by the automated retriever. For example, the verb \textit{yanuy} `to cook' does not appear as a headword in the dictionary, but rather as a regional variant of \textit{wayk'uy} `to cook'.
We also eliminate 
content
from the grammar
that was retrieved because of syncretism,
or mistakes that cascaded from the morphological parser to result in irrelevant retrievals. 

We manually parse each source sentence to only retrieve and include relevant information in the prompt context. All content in the revised prompts is sourced from the same material available to the automated retriever systems, and we do not add any additional information or use supplemental materials of any sort to create the revised prompts.

\subsection{Models}
We experiment with three proprietary models, GPT-3.5 Turbo \citep[\texttt{gpt-3.5-turbo-0125},][]{brown2020language}, GPT-4o \citep[\texttt{gpt-4o},][]{achiam2023gpt4}, and Gemini 1.5 Pro \citep[\texttt{gemini-1.5-pro},][]{reid2024gemini}, and one open-source model, Llama 3 \citep[\texttt{llama-3-8b-instruct},][]{llama3modelcard}. We use the pretrained models with their default settings, and do not adjust hyperparameters or conduct any finetuning as part of our experiments.

\subsection{Evaluation}
\label{sec:eval}
We conduct both automatic and human evaluations
to identify trends in model errors and outputs in the various experimental conditions.
We calculate BLEURT and BLEU scores as automatic metrics, and report mean BLEURT scores across items as the primary quantitative measure of translation quality for each of the conditions and models. We also use an adapted MQM schema 
to conduct qualitative human evaluation of the outputs of GPT-3.5 and GPT-4o for all prompts using automatic retrieval.

Each item selected for human evaluation is annotated by at least one of the authors by comparing the model's output to the source text and reference translation.
We refer to the
complete MQM typology to design our own four-dimensional framework of commonly attested errors in LLM-MT,
each with a defined set of specific subtypes. 
Precise definitions and examples for all error categories and subtypes may be found in Appendix \ref{app:errortypes}.

Many of the categories in our schema are defined as in the core MQM framework. However, to capture some of the key behaviors reported in previous studies on LLM-MT and to evaluate the effects of prompt type on model outputs, we make the following adjustments.
First, we utilize the Addition and Omission errors defined as Accuracy subtypes in the original MQM typology, but distinguish these from three additional subtypes: Substitution - Incorrect Subject, Substitution - Incorrect Tense/Aspect/Modality (TAM), and Substitution - Other. This is intended to capture LLM translations that differ from the source in terms of discrete lexical material or case, person, number, and/or TAM markings while otherwise maintaining the lexical and  structural content needed to appropriately translate the source text. Although they are not the only grammatical phenomena that may be similarly misrendered, we select subject and TAM markers for analysis as they are straightforward to identify and give a good indication of how well the LLMs cope with more abstract information about the meanings of functional morphemes.


Rather than including Mistranslation and MT Hallucination as Accuracy Error subtypes as in the original MQM typology, we define a separate Non-Translation category with three possible subtypes: Complete Mistranslation, Mistranslation with Lexical Correspondences, and Refusal. The third dimension of our typology, Model Error, was ultimately not used to classify any output in this study, but characterizes more generic model ``misbehavior'' such as failing to follow instructions, producing garbled text, or inappropriately generating content in the source language. Finally, Target Errors identify outputs that are ungrammatical, stylistically inappropriate, or semantically incoherent in the target language, regardless of their accuracy.

Detailed annotation guidelines were drafted and agreed upon to encourage consistency across annotators and experimental items. Annotators are instructed to identify and tag up to three specific errors for each translation output, with the exception of Target Errors, which do not count towards the three-error maximum. Each model output is also tagged for quality along a four-point scale as defined in Table \ref{table:quality}. 

Before proceeding with annotation over the larger dataset, both annotators also completed a test evaluation of the same 12 experimental items (96 sentences total) to assess inter-annotator agreement. Statistical measures ($\kappa$ = 0.72 for quality judgments, $\alpha$ = 0.55 for error categories) indicated some discrepancies in annotator judgments, especially for categories, since determining the three most important errors is especially subjective. These were identified and discussed, and agreement was ultimately deemed sufficient to proceed.

\section{Results}

\begin{table}
\centering
    \begin{tabular}{p{1.1cm}cccc}
       & GPT3.5 & GPT4o & Gem. & Lla3\\
       \hline
\textsc{base}	&	0.19	&	0.66	&	0.56	&	0.15\\
\textsc{corpus}	&	0.27	&	0.59	&	0.49	&	0.19\\
\textsc{gram}	&	0.23	&	0.56	&	0.55	&	0.17\\
\rowcolor{CornflowerBlue} \textsc{morph}	&	0.44	&	0.54	&	0.61	&	0.39\\
\textsc{c+g} &	0.26	&	0.59	&	0.54	&	0.21\\
\rowcolor{CornflowerBlue} \textsc{c+m}	&	0.44	&	0.59	&	0.59	&	0.36\\
\rowcolor{CornflowerBlue} \textsc{g+m}	&	0.41	&	0.53	&	0.61	&	0.39\\
\rowcolor{CornflowerBlue} \textsc{c+g+m}	&	0.43	&	0.57	&	0.61	&	0.15\\
    \end{tabular}
    \caption{Mean BLEURT scores by LLM and prompt type. Shaded rows include morpheme contexts.}
    \label{tab:bleurtscores}
\end{table}

\begin{table}
    \centering
    \begin{tabular}{lcc}
LLM & GPT-3.5 & GPT-4o \\
\hline
\textsc{baseline}    &   21      & 108     \\
\textsc{corpus} &   43      & 101     \\
\textsc{grammar}    &   33  & 99     \\
\rowcolor{CornflowerBlue} \textsc{morph} &   79      & 102     \\
\textsc{c+g}  &   41  & 101     \\
\rowcolor{CornflowerBlue} \textsc{c+m}   &   75  & 110     \\
\rowcolor{CornflowerBlue} \textsc{g+m}  &   68  & 100    \\
\rowcolor{CornflowerBlue} \textsc{c+g+m}   &   77  & 109\\
\end{tabular}
\caption{Human-annotated quality ratings summarized as $3\times{high} + 2\times{med} + {low}$. Shaded rows include morpheme contexts.}
\label{tab:qualityscores}
\end{table}

\begin{table}
\centering
    \begin{tabular}{lcccc}
      &  GPT3.5 & GPT4 & Gem. & Lla3\\
      \hline \hline
\textsc{g-auto} & 0.23 & 0.56 & 0.55 & 0.17\\
\textsc{g-man} & 0.24 & 0.58 & 0.54 & 0.15\\
\hline
\textsc{m-auto} & 0.44 & 0.54 & 0.61 & 0.39\\
\textsc{m-man} & 0.56 & 0.63 & 0.66 & 0.49\\
\hline
\textsc{cgm-auto} & 0.43 & 0.57 & 0.61 & 0.15\\
\textsc{cgm-man} & 0.54 & 0.63 & 0.63 & 0.26\\
    \end{tabular}
    \caption{Comparison of mean BLEURT scores for automatic versus manual retrieval of material in \textsc{grammar}, \textsc{morph}, and \textsc{corpus-grammar-morph} prompts.}
    \label{tab:manualscores}
\end{table}

\subsection{Quality Metrics}

We present BLEURT scores for prompts generated using automated retrieval in Table \ref{tab:bleurtscores} and summarize human quality judgments for GPT-3.5 and GPT-4o with automated retrieval in Table \ref{tab:qualityscores}. The complete distribution of BLEURT, BLEU, and human-annotated quality ratings for all of our experiments is provided in Appendix \ref{app:qualityscores}. We find clear effects of LLM, prompt type, and retrieval method, as well as interactions among all three factors. 

Comparing across models, we find that Gemini and GPT-4o outperform Llama 3 and GPT-3.5 for every prompt type. This gap is highest for the least informative prompts, indicating that the Llama 3 and GPT-3.5 base models have relatively poor coverage of S. Quechua, while
GPT-4o and Gemini have much better coverage. The effect is evident in both automatic and human quality evaluations.

Effects of prompt type are mediated by the quality of the pretrained model. Llama 3 and GPT-3.5 show a clear improvement in quality when \textsc{morph} information is included in the prompt. Gemini also improves when this information is added, but to a lesser extent. GPT-4o, on the other hand, performs best in response to the \textsc{baseline} (zero-shot) prompts, which attain the highest BLEURT scores across all models, prompt types, and retrieval methods evaluated in this study. In other words, providing additional information in the prompt's context actually \textit{degrades} GPT-4o's ability to translate from S. Quechua to Spanish in all experimental conditions.

\subsection{Effects of Automated Retrieval}
To highlight the effects of automated retrieval on model output, we present BLEURT scores for a selection of prompt types and all four models in Table \ref{tab:manualscores} (full scores may be found in Appendix \ref{app:qualityscores}). 
The effect of manual retrieval for \textsc{morph} information is positive for all models, although this gap is smallest for Gemini (probably because its performance for these prompts is already highest). The effect for \textsc{grammar} prompts is either minor or negative.

\subsection{Human Analysis of Translation Errors}
\label{sec:specificerrors}

The most common error type identified by the annotators is Substitution - Other, which includes a diverse assortment of lexical and phrasal incongruencies of varying degrees of severity. These are largely item-specific and therefore hard to characterize as a group. Using the error categories described in Section \ref{sec:eval}, we instead identify three more clearly interpretable phenomena and provide a detailed discussion of each in the following sections. We present counts for selected prompt types in Table \ref{tab:finegrained}, with examples in Appendix \ref{tab:errorexe} and counts for all errors in Appendix \ref{app:errorcounts}.  


\begin{table*}
\centering
\begin{tabular}{llccc}
& & \textsc{base} & \textsc{morph} & \textsc{c+g+m} \\
\hline \hline
\multirow{2}{5cm}{Mistranslation: complete + lexical correspondence} & GPT-3.5 & 45 & 11 & 12 \\
& GPT-4o & 4 & 6 & 4\\
\hline
\multirow{2}{5cm}{Target Fluency: grammar + coherence + style} & GPT-3.5 & 0 & 14 & 10 \\
& GPT-4o & 3 & 13 & 9 \\
\hline
\multirow{2}{5cm}{Grammatical Divergence: subject + TAM} & GPT-3.5 & 0 & 24 & 31\\
& GPT-4o & 17 & 13 & 11 
\end{tabular}
\caption{Counts of human-annotated error types (per 50 sentences) by LLM and prompt type.}
\label{tab:finegrained}
\end{table*}

\subsection{Mistranslations}
\label{sub:mistrans}

Outright mistranslations are most common for GPT-3.5, making up 30 of the 50 responses in the \textsc{baseline} condition.
We also consider outputs that retain only minimal traces of the source content, which we label as Mistranslations with Lexical Correspondence. Approximately 1/3 of the 637 total errors tagged across all prompt types for GPT-3.5 are mistranslations of either type, roughly split between complete mistranslations and those with lexical correspondence (15.07\% and 18.37\%, respectively, of all errors tagged for GPT-3.5). 

As reported in previous work, adding morpheme- and word-level translations to the prompt greatly reduces the rate of this kind of response. GPT-4o also produces drastically fewer mistranslations compared to its predecessor. However, it is notable that both models produce at least one mistranslation for each prompt type. 
In general, complete mistranslations are in fluent Spanish and contain no overt indications that something has been misrepresented. We return to the ethical implications of these errors in the \nameref{sec:discussion}.

We also note that many of the items tagged as Mistranslation with Lexical Correspondence show correspondence only for words that were already in Spanish in the source text. For example, 
some sentences contain Spanish loan names for the days of the week.
While some of these errors are produced in deceptively fluent Spanish, we find many to be accompanied by semantic incoherence or ungrammaticality in the output. We discuss such target language fluency errors in the following section.

\subsection{Target Fluency}

Target Fluency errors occur when the output is not grammatical, coherent, or stylistically appropriate -- for instance, if an output contains a nonsensical repetition or a verb with missing arguments. Outputs of this type bear a strong similarity to human ``translationese'' in that structural features of the source language may surface in the translation at the expense of naturalness \cite{koppel-ordan-2011-translationese,freitag-etal-2019-ape}. Both GPT-3.5 and GPT-4o tend to produce more such outputs when the prompt is more informative  -- 10 to 20\% of the time (5-10 instances per 50) in prompts with morpheme translations. 


\subsection{Grammatical Divergence}

We group misrendered verbal subjects and tense/aspect/morphology (TAM) markers together as Grammatical Divergence errors. Such errors are distinct from the Target Fluency errors described in the previous section--- the Spanish output 
is grammatical, but fails to accurately reflect the content from the source.
TAM divergences are much more prevalent than divergences in subject; for instance, only one of GPT-4o's 13 Grammatical Divergence errors in the \textsc{morph} condition misrender the subject marker.

Grammatical Divergence errors are annotated only for sentences that are not mistranslated outright, so GPT-3.5 produces none of these in the \textsc{baseline} condition. For more informative prompts, it is clear that GPT-4o is better than GPT-3.5 at translating both functional and lexical meanings. However, a relatively large number of sentences (over 20\%) still contain such an error even with the highest performing model and prompt type. The relatively small drop in error between different prompt types for GPT-4o suggests that neither the corpus-based usage examples nor example paradigms and descriptions from the grammar document can fully prevent this type of error. 

\section{Discussion}
\label{sec:discussion}
We observe large differences between LLMs, both in terms of the overall quality of their generated translations as well as the effects of prompt type on their outputs. GPT-4o and Gemini, which have the highest baseline scores, benefit least from additional information--- BLEURT scores actually decrease when \textsc{corpus} and \textsc{grammar} information is included. This occurs even with manually curated prompts, suggesting it is not an effect of including irrelevant material. Nonetheless, the baseline results do not represent a ceiling on quality, since both models still produce errors in the \textsc{baseline} condition (GPT-4o produces 10 \textsc{low}-quality translations in our set of 50). These results suggest that even relevant grammar explanations, when written in prose with examples, do little to help the newest generation of LLMs to translate a low-resource language such as Southern Quechua.

Although GPT-4o and Gemini results are similar in many ways, we do find evidence for differences in their in-context learning abilities. Baseline prompts and the GPT-4o model produce the highest BLEURT scores across the dataset, but these outputs still show a number of errors characteristic of LLMs, particularly lexical substitution errors that are not necessarily corrected with the inclusion of more context. In contrast, Gemini, which has near-comparable performance across prompt types, shows an increase in scores when prompts include \textsc{morph} information, regardless of retrieval type, suggesting a greater ability to identify and utilize relevant word- and morph-level translations in the prompt's context. Previous work suggests that newer builds of GPT-4 are less capable of following instructions \cite{chen2023analyzing}; such differences may be masked by the effects of pretraining when automatically evaluating translations. This suggests that researchers should continue to carefully select and compare among different LLMs when experimenting with retrieval-based translation.

\subsection{Language-Specific Effects}
\label{sec:lang-effects}
We identify a number of translation errors of varying types that appear to be due to language-specific factors such as those discussed in Section \ref{sec:morphology}.
For example, we find an effect of polysemous lexical items
for all prompt types in the outputs of both models on which we conduct human evaluation. 

In the most straightforward cases, the model incorrectly generates an alternate sense of the word that is inappropriate given the content of the source sentence.
We also find a number of instances in which the presence of such ambiguity has a cascading effect on lexical selection in the rest of the model's output. For example, when co-occuring with \textit{miski} in the source text, the word \textit{lawa} `soup' is translated at times as \textit{mazamorra}, a sweet porridge or pudding, \textit{crema} `cream', \textit{miel} `honey', \textit{golosina} `candy', or \textit{dulces}, the nominal form of \textit{dulce} meaning `candy' or `sweets.' 

It may be possible to moderate such effects with additional refinement of the database structure and retrieval methods, which we leave for future work.

\subsection{Ethical Concerns}
\label{sec:ethics}

Both our work and much of the previous work in this paradigm is motivated by the desire to close the ``NLP Gap'' among researchers, community members, and software developers interested in low-resource language technologies. Machine translation is listed as a welcome topic of research by some (though not all) members of American Indigenous communities \cite{mager-etal-2023-ethical}, and is potentially an important tool for language learners \cite{jolley2022thirty}. Even an imperfect translation system might be a useful tool for users with a clear understanding of its limitations. However, the systems evaluated in this work have two problematic tendencies that limit their potential for deployment in real community settings.

First, unfaithful translations often tend to be highly fluent (Section \ref{sub:mistrans}). While fluency ratings for older MT systems correlate well with accuracy scores, and have even been used as a proxy for overall translation quality \cite{gamon2005sentence,estrella-etal-2007-new}, this correlation is reversed for our systems. LLMs are well-known for making false statements that seem plausible and authoritative \cite{bickmore2018,dinan2021anticipating}; this could be particularly problematic when they project illusions of expertise at the expense of an already marginalized group.

Second, some mistranslations identified in our study appear to draw on stereotypes of Indigenous groups (Appendix \ref{app:stereotypes}). These are most apparent for the \textsc{baseline} system and GPT-3.5, but also (less frequently) occur with more informative prompts and better LLMs. Stereotypical sentences can involve flowery language with an emphasis on tradition or connectedness to nature \cite{erhart2019descriptive}, as well as the unprompted addition of Indigenous Andean cultural customs and products (e.g., traditional medicine and chicha) to translations that are otherwise faithful to the source text. The overall effect is to exoticize Southern Quechua speakers and writers in ways that the original sentences do not. Similar stereotypes have also been noted in LLM-generated responses to open-ended prompts \cite{cheng-etal-2023-marked, delgato-toxtli-2023-stereotypes, shieh2024laissezfaire}.

While we prompt models to output only the translation for evaluation purposes, models may have some capacity to explain or qualify their translations and give
reminders for responsible use of the technology. Should a retrieval-based translation system ever be deployed in a real-world setting for language learning, its developers should maximize transparency by presenting the content of any retrieved information and its source to the user along with the translation, reminding users directly of potential inaccuracies,
and offering vetted resources for additional fact-checking when available.

\section{Conclusion}

Our results suggest a number of key limitations and concerns regarding the use of LLMs in a low-resource MT context, and have greater implications for our understanding of the seemingly ``humanlike'' conceptual, analytical, and in-context learning abilities of LLMs.

For the majority of the world's languages and their speakers, powering and supplying LLMs with enough pretraining data to overcome their limitations is not feasible.
We therefore offer the following suggestions to those looking to develop low-resource LLM-MT: (1) improve data structures and methods for interacting with a language-specific database for retrieval-aided generation, (2) continue analysis of the mechanisms driving in-context learning in LLMs, for example by comparing ICL to the effects of finetuning \cite{dai2023gpt}, and (3) experiment with prompt structures and techniques, for example by altering the order of information \cite{liuetal-2024-lost} or by iteratively prompting the model to guide its reasoning towards a suitable translation \cite{wang2022iteratively}.

Finally, we wish to emphasize the continued risks of prematurely deploying this or similar methods in any low-resource language community, particularly given the vulnerability and disproportionate lack of resources many such communities face in domains where these technologies would likely be used. As AI research continues to rapidly develop, we urge those conducting it to increase community engagement, amplify the voices of those traditionally at a disadvantage, and collaboratively develop research infrastructures that may lessen the NLP Gap \citep{brinklow2021indigenous}. While there's still much to be done before low-resource LLM-MT may be safely implemented, we believe such a tool has the potential to empower speakers of any variety, including nonstandard varieties of high-resource languages such as English, to develop technologies that reflect their preferences and serve their unique needs.

\section{Limitations}
\label{sec:limits}
Limitations on the scope and replicability of this work may be attributed to one or more characteristics of the data and models used in this study, in addition to limitations inherent to the respective identities of its authors. First, the automatic metrics (i.e., BLEURT and BLEU scores) that we report are limited in their statistical validity. We have conducted some constrained tests to explore potential variance in scores, but expenses associated with text generation using proprietary models such as those developed by OpenAI and Google on a larger dataset may be prohibitive. This is compounded by the widely-acknowledged ``black box'' nature of the models powering both LLMs and BLEURT, as well as an increasing opacity with respect to the exact content and methods used to pretrain modern state of the art LLMs. For this reason, we focus our discussion on those results that show clear trends in both the quantitative and human evaluations we conduct.

There are also some constraints on our study and its methodology that are largely tied to linguistic factors, such as variation in orthography (and the need for digitized text-based resources as a prerequisite) as well as the lexical and grammatical variation that may be found in all languages, particularly the low-resource varieties we wish to support. We discuss some of these factors in Sections \ref{sec:morphology} and \ref{sec:lang-effects}. Our results suggest it may be possible to guide the outputs of LLMs towards the specific usage conventions of a given community, but this is itself limited by the content of the materials used to develop the database from which prompt contexts are retrieved. 

Neither of the authors is a native speaker of any Quechua or Spanish varieties, and only one is a student of these languages and has relationships to Quechua speakers and communities. While we have strived to be consistent in the Quechua and Spanish varieties used in our study (both the dictionary and grammar materials were provided by the same instructor who shared and proofread the 50 sentence pairs we use, and we select a morphological parser and corpora intended for use with Southern Quechua), variation is widespread among and within Quechua-speaking communities, and we do not have access to a dictionary, grammar, morphological parser, and corpus developed by a unified and consistent set of authors. 
Future work should continue to explore ways to faithfully represent the diversity of linguistic conventions employed by communities interested in developing such technologies.

We acknowledge, as well, limitations that arise from the size of our dataset and database and the methods used to curate them. The 50 sentence pairs we use were selected to highlight a range of specific grammatical phenomena, not all of which were well represented in our database, and differ in their structural complexity. We are grateful for the guidance provided by the Quechua instructor whose lessons were a source for such examples and proofread the sentences before their inclusion in our experiments, but are limited by our status as non-native speakers. Human evaluation of model outputs was partially conducted using machine-translated English texts as references, but all annotations were inspected by the Spanish- and Quechua-speaking author who removed a small number of evaluations that reflected linguistic discrepancies between Quechua, Spanish, and English or inaccuracies in the machine-translated English.

\section{Ethics Statement}

We consulted the first author's Quechua instructor, Prof. Carmen Cazorla Zen, who gave us permission to use the sentences from the notes in this project and verified their accuracy.
We cite the Quechua dictionary and grammar materials used to provide prompt information, and believe that our use of these materials
is consonant with their original purpose.
However, we do not distribute machine-readable versions of them as a contribution of this project, since this would violate the rights of the publisher. These materials were developed for use as pedagogical resources by institutions affiliated with the governments of Cuzco, Peru and Apurímac, Peru, respectively. Their authors were not contacted or consulted as part of the project. 

We wish to acknowledge the delicate issue of academic \textit{extractiveness} and its harmful impact on Indigenous and minority language communities and speakers. We are also aware of some of the controversial ideologies and policies associated with Qheswa Simi Hamut'ana Kuraq Suntur, the government-afilliated institution who published the dictionary we use in this study, and the potentially negative effects of government-sponsored linguistic standardization more broadly (see, e.g., \citeauthor{coronelmolina2008} (\citeyear{coronelmolina2008}) for an analysis of the effects of the institution's ideologies on revitalization efforts in Peru). We do not endorse such policies, and have sought to avoid representing the diversity of Southern Quechua-speaking communities as a monolith.
Instead, we hope our continued efforts to improve methods for low-resource translation will empower speakers of Southern Quechua and other Indigenous and minority languages to develop language technologies capable of representing their own community's unique language variety to serve the unique needs of its speakers.

There are numerous ethical issues related to the training and use of LLMs, such as labor issues and energy costs. While these issues are inextricable from the methods used in this project, we believe the potential impact of making low-resource translation viable and accessible to minority language communities who want them (our primary goal in this line of research) outweighs the problems inherent in using LLMs at all.
We discuss the potential risks of deploying systems like the ones described here further in Section \ref{sec:ethics} of the main text. 
%




\section*{Acknowledgments}
We thank Prof. Carmen Cazorla Zen, Professor of Quechua, for her help curating the data used in this study and for deepening our understanding of Southern Quechua and its speakers. We also thank Prof. Elvia Andía Grágeda, Professor of Quechua, for her instruction and advice, and the OSU Linguistics department for their feedback on a preliminary presentation of the work.

\bibliography{custom}

\newpage
\appendix


\section{Example Errors}
\label{tab:errorexe}
\noindent \textit{The following section provides examples of errors analyzed in Section \ref{sec:specificerrors}, one error per type.}\\

\hrule
\vspace{1pt}
\hrule
\vspace{5pt}

\noindent \\ \textbf{Mistranslation: Complete Mistranslation}\\
\hrule
\vspace{1pt}
\hrule
\vspace{10pt}

\noindent \textbf{Model:} GPT-3.5 - \textsc{baseline} - \textsc{auto}\\


\noindent \textbf{Source:} qamqa taytaykipa munasqan lawata yanurqanki \\

\noindent \textbf{Gloss:}
\nonumber \gll qam-qa tayta-yki-pa muna-sqa-n lawa-ta yanu-rqa-nki \\
you-\textsc{top} father-\textsc{2.poss}-\textsc{gen} like-\textsc{subr}-\textsc{3sg} soup-\textsc{acc} cook-\textsc{pst}-\textsc{2sg} \\
\vspace{5pt}

\noindent \textbf{Reference:} tú cocinaste la sopita que le gusta a tu papá \\
\noindent \textit{you cooked the soup that your dad likes}\\

\noindent \textbf{Output:} yo te quiero mucho y mi amor por ti nunca se acabará\\
\noindent \textit{I love you very much and my love for you will never end} \\

\hrule
\vspace{1pt}
\hrule
\vspace{10pt}

\noindent \textbf{Mistranslation: Lexical Correspondence}\\

\hrule
\vspace{1pt}
\hrule
\vspace{10pt}

\noindent \textbf{Model:} GPT-4o - \textsc{c+g+m} - \textsc{auto} \\


\noindent \textbf{Source:} ñuqaqa illani runakunawan rimayta munaptiy\\  

\noindent \textbf{Gloss:}

\gll ñuqa-qa illa-ni runa-kuna-wan rima-y-ta muna-pti-y\\  
I-\textsc{top} travel-\textsc{1sg} person-\textsc{pl}-with speak-\textsc{acc} like-\textsc{subr}-\textsc{1sg}\\
\vspace{5pt}

\noindent \textbf{Reference:} yo viajo porque me gustar hablar con las personas \\
\noindent \textit{I travel because I like to talk to people} \\

\noindent \textbf{Output:} yo quiero hablar con las personas cuando amanece\\ 
\textit{I want to talk to people when the sun rises}\\

\hrule
\vspace{1pt}
\hrule
\vspace{10pt}
\noindent \textbf{Target Fluency: Coherence }\\
\textit{Unresolvable pronoun}\\

\hrule
\vspace{1pt}
\hrule
\vspace{10pt}

\noindent \textbf{Model:} GPT-3.5 -  \textsc{c+g+m} -\textsc{auto} \\


\noindent \textbf{Source:} qamqa yachayta tukuptiyki, puñunki\\  

\noindent \textbf{Gloss:}

\gll qam-qa yacha-y-ta tuku-pti-yki, puñu-nki\\
you-\textsc{top} study-\textsc{inf}-\textsc{acc} finish-\textsc{subr}-\textsc{2sg}, sleep-\textsc{2sg}\\
\vspace{5pt}

\noindent \textbf{Reference:} cuando termines de estudiar, tu duermes\\
\noindent \textit{when you finish studying, you sleep} \\

\noindent \textbf{Output:} acabaré de aprender lo que tú duermes\\ \textit{I will finish learning what you sleep}\\

\hrule
\vspace{1pt}
\hrule
\vspace{10pt}

\noindent \textbf{Target Fluency: Grammar}\\
\textit{Missing causative and possessive}
\vspace{10pt}
\hrule
\vspace{1pt}
\hrule
\vspace{10pt}

\noindent \textbf{Model:} GPT-4o - \textsc{corpus} - \textsc{auto}\\


\noindent \textbf{Source:} ñuqaqa ñañayman wasita pichachini\\

\noindent \textbf{Gloss:}

\gll ñuqa-qa ñaña-y-man wasi-ta picha-chi-ni\\
I-\textsc{top} sister-\textsc{1sg.poss}-\textsc{ill} house-\textsc{acc} clean-\textsc{caus}-\textsc{1sg}\\
\vspace{5pt}

\noindent \textbf{Reference:} yo le hago limpiar la casa a mi hermana \\
\noindent \textit{I make my sister clean the house}\\

\noindent \textbf{Output:} yo hermana casa limpio\\
\textit{I sister house clean} \\


\hrule
\vspace{1pt}
\hrule
\vspace{10pt}

\noindent \textbf{Grammatical Divergence: TAM}\\
\textit{Tense}\\

\hrule
\vspace{1pt}
\hrule
\vspace{10pt}

\noindent \textbf{Model:} GPT-3.5 -  \textsc{c+g+m} - \textsc{auto}\\


\noindent \textbf{Source:} qam allinta tusunki\\  

\noindent \textbf{Gloss:}

\gll qam allin-ta tusu-nki\\
you good-\textsc{acc} dance-\textsc{2sg}\\
\vspace{5pt}

\noindent \textbf{Reference:} tu bailas bien \\
\noindent \textit{you dance well} \\

\noindent \textbf{Output:} tú bailarás bien\\
\textit{you will dance well} \\
\newpage
\hrule
\vspace{1pt}
\hrule
\vspace{10pt}

\noindent \textbf{Grammatical Divergence: Subject}\\
\textit{`you' to `I'}\\
\hrule
\vspace{1pt}
\hrule
\vspace{10pt}

\noindent \textbf{Model:} GPT-3.5 - \textsc{morph} - \textsc{auto}\\


\noindent \textbf{Source:} qamqa mamaykipa wasinta yanuq rinki\\  

\noindent \textbf{Gloss:}

\gll qam-qa mama-yki-pa wasi-n-ta yanu-q ri-nki\\
you-\textsc{top} mother-\textsc{2.poss}-\textsc{gen} house-\textsc{3sg.poss}-\textsc{acc} cook-\textsc{agt} go-\textsc{2sg}\\
\vspace{5pt}

\noindent \textbf{Reference:} tú irás a cocinar a la casa de tu mamá\\
\noindent \textit{you will go to cook at your mother's house}\\

\noindent \textbf{Output:} voy a cocinar en la casa de tu madre \\ \textit{I'm going to cook at your mother's house}\\

\hrule
\vspace{1pt}
\hrule
\vspace{10pt}

\noindent \textbf{Grammatical Divergence: Subject}\\
\textit{`you' to `the man'}
\vspace{10pt}
\hrule
\vspace{1pt}
\hrule
\vspace{10pt}

\noindent \textbf{Model:} GPT-4o - \textsc{c+g+m} - \textsc{auto}\\


\noindent \textbf{Source:} qamqa misita qawaspa qari warata awachkarqanki\\  

\noindent \textbf{Gloss:}

\gll qam-qa misi-ta qawa-spa qari wara-ta awa-chka-rqa-nki\\
you-\textsc{top} cat-\textsc{acc} watch-\textsc{subr} male pants-\textsc{acc} knit-\textsc{prog}-\textsc{pst}-\textsc{2sg}\\
\vspace{5pt}

\noindent \textbf{Reference:} tú estabas tejiendo un pantalón de varón mirando al gato\\
\noindent \textit{you were knitting a pair of men's pants looking at the cat}\\

\noindent \textbf{Output:} tú mirabas al gato mientras el hombre tejía el pantalón\\
\textit{you looked at the cat while the man knitted the pants}  \\

\hrule
\vspace{1pt}
\hrule
\vspace{10pt}

\noindent \textbf{Grammatical Divergence: TAM}\\
\textit{Missing conditional}
\vspace{10pt}
\hrule
\vspace{1pt}
\hrule
\vspace{10pt}

\noindent \textbf{Model:} GPT-4o - \textsc{morph} - \textsc{auto}\\


\noindent \textbf{Source:} qam runasimita yachapayaptiyki, allintam rimanki\\  

\noindent \textbf{Gloss:}

\gll qam runa-simi-ta yacha-paya-pti-yki, allin-ta-m rima-nki\\
you people-mouth-\textsc{acc} learn-\textsc{rep}-\textsc{cond}-\textsc{2sg}, good-\textsc{acc}-\textsc{asrt} speak-\textsc{2sg}\\
\vspace{5pt}

\noindent \textbf{Reference:} si practicas el quechua lo hablarás bien\\
\noindent \textit{if you practice Quechua you will speak it well}\\

\noindent \textbf{Output:} tú ya sabías el runasimi, hablas muy bien \\
\textit{you already knew runasimi, you speak very well}\\


\newpage
\section{Unprompted Stereotypes}
\label{app:stereotypes}
\textit{The following section provides examples of errors involving cultural stereotypes as discussed in Section \ref{sec:ethics}.} \\

\hrule
\vspace{1pt}
\hrule
\vspace{10pt}

\noindent \textbf{Model:} GPT-3.5 - \textsc{baseline} - \textsc{auto}\\

\noindent \textbf{Source:} payqa qatupi ñañanpa rantisqan apita rurarqan\\  

\noindent \textbf{Gloss:}

\gll pay-qa qatu-pi ñaña-n-pa ranti-sqa-n api-ta rura-rqa-n\\
3-\textsc{top} market-\textsc{loc} sister-\textsc{3sg.poss}-\textsc{gen} buy-\textsc{subr}-\textsc{3sg} mazamorra-\textsc{acc} make-\textsc{pst}-\textsc{3sg}\\
\vspace{5pt}

\noindent \textbf{Reference:} ella preparó una mazamorra que compró su hermana en el mercado \\
\noindent \textit{she prepared a mazamorra [pudding] that her sister bought at the market} \\

\noindent \textbf{Output:} el sol brilla en nuestra tierra, calienta y nos da vida\\
\textit{the sun shines on our land, warms us and gives us life} \\

\hrule
\vspace{1pt}
\hrule
\vspace{10pt}

\noindent \textbf{Model:} GPT-3.5 - \textsc{grammar}\\

\noindent \textbf{Source:} paykunaqa yachachiqninkupa rakta medicina liwru apasqanta likanku \\  

\noindent \textbf{Gloss:}

\gll pay-kuna-qa yacha-chi-q-ni-nku-pa rakta medicina liwru apa-sqa-n-ta lika-nku\\
3-\textsc{pl}-\textsc{top} learn-\textsc{caus}-\textsc{agt}-\textsc{ni}-\textsc{3pl.poss}-\textsc{gen} thick medicine book carry-\textsc{subr}-\textsc{3sg}-\textsc{acc} read-\textsc{3pl}\\
\vspace{5pt}

\noindent \textbf{Reference:} ellos leerán el libro grueso de medicina que trajo su profesor\\
\noindent \textit{they will read the thick medical book that their teacher brought} \\

\noindent \textbf{Output:} los médicos de la comunidad nos enseñaron a utilizar plantas medicinales\\ 
\textit{the doctors from the community taught us how to use medicinal plants}\\

\hrule
\vspace{1pt}
\hrule
\vspace{10pt}

\noindent \textbf{Model:} GPT-4o - \textsc{baseline}\\

\noindent \textbf{Source:} qamkunapa taytamamaykichikta watukunaykichik, paykunata purikuq apanaykichik\\  

\noindent \textbf{Gloss:}

\gll qam-kuna-pa tayta-mama-ykichik-ta watuku-na-ykichik, pay-kuna-ta puri-ku-q apa-na-ykichik\\
you-\textsc{pl}-\textsc{gen} father-mother-\textsc{2pl.poss}-\textsc{acc} visit-must-\textsc{2pl}, 3-\textsc{pl}-\textsc{acc} walk-enjoy-\textsc{agt} bring-must-\textsc{2pl}\\
\vspace{5pt}

\noindent \textbf{Reference:} ustedes tienen que visitar a sus padres y llevarlos a pasear\\
\noindent \textit{you have to visit your parents and take them for a walk}\\

\noindent \textbf{Output:} ustedes deben honrar a sus padres, guiarlos en su camino\\
\textit{you must honor your parents, guide them on their path} \\

\hrule
\vspace{1pt}
\hrule
\vspace{10pt}

\noindent \textbf{Model:} Gemini 1.5 - \textsc{baseline}\\

\noindent \textbf{Source:} ñuqaqa upyayta munani\\  

\noindent \textbf{Gloss:}

\gll ñuqa-qa upya-y-ta muna-ni\\
I-\textsc{top} drink-\textsc{inf}-\textsc{acc} want-\textsc{1sg}\\
\vspace{5pt}

\noindent \textbf{Reference:}  yo quiero beber\\
\noindent \textit{I want to drink}\\

\noindent \textbf{Output:} yo quiero beber chicha\\
\textit{I want to drink chicha}\\

\newpage
\begin{table*}[h]
\section{Quality Descriptions}
\label{app:qualitydescriptions}

\centering
\begin{tabular}{|>{\centering\arraybackslash}m{2cm}|>{\RaggedRight\arraybackslash}m{14cm}|}

\hline
\textbf{Quality} & \textbf{Description} \\ \hline
High & Output is an accurate and/or acceptable translation of the source content. \\ \hline
Med & Output contains errors that prevent it from being an acceptable translation, but is generally high in quality otherwise. \\ \hline
Low & Output contains errors that prevent it from being an acceptable translation, with minor correspondences that vaguely identify it as relevant to the source. \\ \hline
None & Output does not appear to be relevant to the source. \\ \hline
\end{tabular}
\caption{Quality Descriptions}
\label{table:quality}
\end{table*}

\begin{table*}[ht]
\section{Annotation Error Typology}
\centering
\label{app:errortypes}
\small
\label{tab:translation-errors}
\begin{tabular}{>{\RaggedRight\arraybackslash}m{2.5cm}>{\RaggedRight\arraybackslash}m{3.1cm}>{\RaggedRight\arraybackslash}m{9cm}}
\textbf{Dimension} & \textbf{Error} & \textbf{ Description} \\ 
\hline
\hline
Accuracy & Addition & Translation includes information not present in the source, but does not result in the displacement of source content. \\ 
\hline
Accuracy & Omission & Translation is missing content from the source. \\
\hline
Accuracy & Substitution - Subject & The translated segment contains content identified as relevant to the source in other spans, but substitutes novel subject markers for those present in the source in the highlighted span; Classify an error as a “substitution” when the error appears to result in both Addition and Omission errors that cannot be distinguished into two distinct spans. \\
\hline
Accuracy & Substitution - TAM & The translated segment contains content identified as relevant to the source in other spans, but substitutes novel TAM for those present in the source in the highlighted span; Classify an error as a “substitution” when the error appears to result in both Addition and Omission errors that cannot be distinguished into two distinct spans. \\
\hline
Accuracy & Substitution - Other & Substitution errors that do not involve mistranslated subject markers or TAM. See above. \\
\hline
Accuracy & Overtranslation & Error occurring in the target content that is inappropriately more specific than the source content. \\
\hline
Accuracy & Undertranslation & Error occurring in the target content that is inappropriately less specific than the source content. \\
\hline
Target Error & Grammar & Other spans in the translated segment may be identified as relevant to the source, but the highlighted span is not grammatical in the target language. \\
\hline
Target Error & Coherence & Other spans in the translated segment may be identified as relevant to the source, but the highlighted span is unnatural or incoherent in the target language. \\
\hline
Target Error & Style/Register & Other spans in the translated segment may be identified as relevant to the source, but the highlighted span is produced in a style or register that is inappropriate given the content. \\
\hline
Non-Translation & Complete Mistranslation & The entire segment is coherent in the target language but the core predicate shows no immediate connection to the reference translation. \\
\hline
Non-Translation & Mistranslation - Lexical Correspondence & The entire segment is coherent in the target language but only minor correspondences to the reference translation may be identified. \\
\hline
Non-Translation & Refusal & Model does not attempt to translate into the target language, e.g., because it "does not understand". \\
\hline
Model error & Garbled & Output does not contain coherent text in the target language. \\
\hline
Model error & ChattyGPT & Output contains translated content, but is wordy, over-explanatory, and/or abruptly truncated. \\
\end{tabular}
\caption{Adapted MQM typology for human error annotation}
\end{table*}

\clearpage\newpage

\section{Example Prompts}
\label{app:exampleprompts}
The following are examples of prompts generated used automated retrieval from the database. English is included in italics for the reader, but was not provided to the models as part of the prompt.\\

\hrule
\vspace{1pt}
\hrule
\vspace{10pt}

\noindent \textbf{\textsc{baseline}}\\

\hrule
\vspace{1pt}
\hrule
\vspace{10pt}

\noindent [TAREA] Traduce la siguiente frase del quechua al español. Responde sólo con la traducción: \\
quechua: qam allinta tusunki\\
español:

\noindent \tikz \draw[dashed] (0,0) -- (\linewidth,0); 

\noindent \textit{[TASK] Translate the following sentence from Quechua to Spanish. Respond only with the translation: \\
Quechua: You dance well \\
Spanish: \\}

\hrule
\vspace{1pt}
\hrule
\vspace{10pt}

\noindent \textbf{\textsc{morphs-only}}\\

\hrule
\vspace{1pt}
\hrule
\vspace{10pt}

\noindent [CONTEXTO]\\
qam: [PrnPers+2sg] \\
allin: bueno [\^DB][NRoot] \\
ta: [+Acc][Cas] \\
tusu: bailar [VRoot][\^DB] \\
nki: [+2sg.Subj][VPers] \\
allin. adj. Bueno (término de aprobación). SINÓN: kusa. EJEM: allin p'unchay, buenos días: allin tuta, buenas noches; allin tutamanta, buena mañana, buenos días; allin inti chinkay, buenas tardes; allin iñiyniyoq, de buena fe, fiel, justo, íntegro: allin nunayoq, de espíritu bueno; allin puriq, de comportamiento bueno; allin puriy, comportamiento bueno; allin rikuy, tratamiento bueno; allin rikuq, el que trata bien; allin ruway, obrar bien, beneficiar; lo que se hace bien, beneficioso; allin ruwaq, el que hace bien; allin yuyay, pensar bien; pensamiento bueno; allin qolqeyoq, poseedor de plata fina; adinerado.\\
ta. s. Gram. Sufijo que desempeña los papeles de artículo y preposición. EJEM: llamata qatiy, arrea la llama; Urkusmanta hamuni, vengo de Urcos. \\

\noindent [TAREA] Traduce la siguiente frase . . . 

\noindent \tikz \draw[dashed] (0,0) -- (\linewidth,0); 

\noindent \textit{[CONTEXT]\\
qam: [PrnPers+2sg] \\
allin: bueno [\^DB][NRoot] \\
ta: [+Acc][Cas] \\
tusu: bailar [VRoot][\^DB] \\
nki: [+2sg.Subj][VPers] \\
allin. adj. Good (term of approval). SYN: kusa. EX: allin p'unchay, good day: allin tuta, good evening; allin tutamanta, good morning, good day; allin inti chinkay, good afternoon; allin iñiyniyoq, good faith, faithful, just, upright: allin nunayoq, in good spirits; allin puriq, well behaved; allin puriy, good behavior; allin rikuy, good treatment; allin rikuq, one who treats others well; allin ruway, to do good, to benefit; one who does good, beneficial; allin ruwaq, one who does good; allin yuyay, think well; good thought; allin qolqeyoq, possessor of fine silver; wealthy.\\
ta. s. Gram. Suffix that plays the roles of article and preposition. EX: llamata qatiy, herd the llama; Urkusmanta hamuni, I come from Urcos. \\}

\noindent \textit{[TASK] Translate the following sentence . . . }\\

\hrule
\vspace{1pt}
\hrule
\vspace{10pt}

\noindent \textbf{\textsc{grammar-only}}\\

\hrule
\vspace{1pt}
\hrule
\vspace{10pt}

\noindent [CONTEXTO]\\
ta: CASO ACUSATIVO. Su marca es –ta, esta es una marca de objeto directo con los verbos que no son de movimiento (quietud). Ejemplo:\\
Quyllur–ta qhawani        Veo una estrella \\
T’anta–ta apay        Lleva pan \\
Ñuqa quylluyta qhawani \\
Pedrucha t’antata rantin \\
En cambio con los verbos de movimiento –ta indica (hacia) que es igual a meta. Ejemplos: \\
Punu–ta rini        Voy a Puno \\
Llaqta-ta risaq        Iré al pueblo \\
Hamawt’anchis Punuta rinqa \\
Llanta umalliq llaqtata richkan \\
nki: FLEXIÓN DE TIEMPO. TIEMPO FUTURO. TIEMPO FUTURO. Los sufijos para cada una de las personas gramaticales son: saq, nki, nqa, sun, saqku, nkichis, nqaku; en singular y plural respectivamente. \\
Ejemplos: \\ 
Puklla-saq        jugaré \\
Puklla-nki        jugarás \\
Puklla-nqa        jugará \\
Puklla-sun        jugaremos \\
Puklla-saqku        jugaremos \\
Puklla-nkichis        Uds. jugarán \\
Puklla-nqaku        ellos jugarán \\

\noindent [TAREA] Traduce la siguiente frase . . . 

\noindent \tikz \draw[dashed] (0,0) -- (\linewidth,0); 

\noindent \textit{\noindent [CONTEXT]\\
ta: ACCUSATIVE CASE. Marked by –ta, this is a direct object marker with verbs that don't indicate movement. Example:\\
Quyllur–ta qhawani        I see a star \\
T’anta–ta apay        Bring bread \\
Ñuqa quylluyta qhawani \\
Pedrucha t’antata rantin \\
On the other hand, with verbs of motion -ta indicates (towards) the same goal Examples: \\
Punu–ta rini        I go to Puno \\
Llaqta-ta risaq        I will go to town \\
Hamawt’anchis Punuta rinqa \\
Llanta umalliq llaqtata richkan \\
nki: TENSE INFLECTION. FUTURE TENSE. FUTURE TENSE. The suffixes for each of the grammatical persons are: saq, nki, nqa, sun, saqku, nkichis, nqaku; in singular and plural respectively. \\
Ejemplos: \\ 
Puklla-saq        jugaré \\
Puklla-nki        jugarás \\
Puklla-nqa        jugará \\
Puklla-sun        jugaremos \\
Puklla-saqku        jugaremos \\
Puklla-nkichis        Uds. jugarán \\
Puklla-nqaku        ellos jugarán \\}

\noindent \textit{[TASK] Translate the following sentence . . . } \\

\hrule
\vspace{1pt}
\hrule
\vspace{10pt}

\noindent \textbf{\textsc{corpus-only}}\\
\hrule
\vspace{1pt}
\hrule
\vspace{10pt}

\noindent [CONTEXTO]\\
quechua: rimanakunapaq wawakunapa rimasqan simi aswan allinta takyachinaraq piwanpas maywanpas mana manchakuspa rimananpaq chaymi qillqanapaqpas ñawichanapaqpas aswan allin kanqa\\
español: para este diálogo saber la lengua que dominan los niños sería importante para que ellos se expresen sin miedo de ahí será que la escritura y la lectura salga de manera óptima\\
quechua: kay tiqsipi sumaq rimanakunapaqa kawsayninchikmi allinta kallpachawanchik runakunahina allinta tiyanapaq chaymi ñuqanchikkqa allinta ñawichayta qillqayta yachananchik ñawpa ayllunchikkuna rurasqankuta maytukunapi tukuy puyñukunapi tiqsi muyu qhawarisqankuta \\
español: para vivir en armonía tenemos que conocer bien nuestra forma de vivir y luego escribir leer tambien a valorar lo que nos dejaron nuestros antecesores en cada visión sobre el mundo \\
quechua: winsislawcha chayarqamuptinsi tuparquspanku allinta qatunakusqanku suwakuypi purinankupaq \\
español: cuando había llegado wenseslau y a su encuentro se habían reforzarón para andar a robar \\

\noindent [TAREA] Traduce la siguiente frase . . .

\noindent \tikz \draw[dashed] (0,0) -- (\linewidth,0); 

\noindent \textit{[CONTEXT]\\
quechua: rimanakunapaq wawakunapa rimasqan simi aswan allinta takyachinaraq piwanpas maywanpas mana manchakuspa rimananpaq chaymi qillqanapaqpas ñawichanapaqpas aswan allin kanqa\\
Spanish: For this dialogue, knowing the language that the children speak would be important for them to express themselves without fear, and that is why writing and reading will be optimal.\\
quechua: kay tiqsipi sumaq rimanakunapaqa kawsayninchikmi allinta kallpachawanchik runakunahina allinta tiyanapaq chaymi ñuqanchikkqa allinta ñawichayta qillqayta yachananchik ñawpa ayllunchikkuna rurasqankuta maytukunapi tukuy puyñukunapi tiqsi muyu qhawarisqankuta \\
Spanish:To live in harmony we have to know our way of life well and then write and read to also value what our ancestors left us in each vision of the world.\\
quechua: winsislawcha chayarqamuptinsi tuparquspanku allinta qatunakusqanku suwakuypi purinankupaq \\
español: cuando había llegado wenseslau y a su encuentro se habían reforzarón para andar a robar}\\

\noindent \textit{[TASK] Translate the following sentence . . . } \\

\begin{table*}[h!]
\section{Full Quality Scores}
\label{app:qualityscores}

This section contains tables showing all automatic and human-annotated quality scores for each of our experiments. Table \ref{tab:fullbleurt} contains the full set of BLEURT scores summarized in Tables \ref{tab:bleurtscores} and \ref{tab:manualscores} of the main text. Table \ref{tab:bleu} shows the corresponding BLEU scores for the same experiments. Table \ref{tab:fullqualitygpt3} and Table \ref{tab:fullqualitygpt4} contain the full set of the human-annotated scores summarized in Table \ref{tab:manualscores}.\\

\centering
    \begin{tabular}{lcccccccc}
 & \multicolumn{2}{c}{\textbf{GPT-3.5}} & \multicolumn{2}{c}{\textbf{GPT-4o}} & \multicolumn{2}{c}{\textbf{Gemini-1.5}} & \multicolumn{2}{c}{\textbf{Llama 3}} \\
\hline
 & auto & manual & auto & manual & auto & manual & auto & manual\\
 \hline \hline
\textsc{baseline}	&	0.19	&	0.22	&	0.66	&	0.66	&	0.56	&	0.57	&	0.15	&	0.16\\
\textsc{corpus-only}	&	0.27	&	0.29	&	0.59	&	0.61	&	0.49	&	0.47	&	0.19	&	0.18\\
\textsc{grammar-only}	&	0.23	&	0.24	&	0.56	&	0.58	&	0.55	&	0.54	&	0.17	&	0.15\\
\textsc{morph-only}	&	0.44	&	0.56	&	0.54	&	0.63	&	0.61	&	0.66	&	0.39	&	0.49\\
	\textsc{corpus-grammar}	&	0.26	&	0.28	&	0.59	&	0.59	&	0.54	&	0.53	&	0.21	&	0.21\\
	\textsc{corpus-morph}	&	0.44	&	0.52	&	0.59	&	0.64	&	0.59	&	0.64	&	0.36	&	0.38\\
\textsc{grammar-morph}	&	0.41	&	0.54	&	0.53	&	0.61	&	0.61	&	0.64	&	0.39	&	0.37\\
	\textsc{corpus-grammar-morph}	&	0.43	&	0.54	&	0.57	&	0.63	&	0.61	&	0.63	&	0.15	&	0.26
    \end{tabular}
    \caption{BLEURT scores for all LLMs and prompt types.}
    \label{tab:fullbleurt}
\end{table*}

\centering
\begin{table*}[h!]
\begin{tabular}{lcccccccc}
 & \multicolumn{2}{c}{\textbf{GPT-3.5}} & \multicolumn{2}{c}{\textbf{GPT-4o}} & \multicolumn{2}{c}{\textbf{Gemini-1.5 Pro}} & \multicolumn{2}{c}{\textbf{Llama 3 8B}} \\
\hline
 & auto & manual & auto & manual & auto & manual & auto & manual\\
 \hline \hline
\textsc{baseline}	&	0.01	&	0.02	&	0.19	&	0.18	&	0.12	&	0.14	&	0.00	&	0.00\\
\textsc{corpus-only}	&	0.02	&	0.02	&	0.16	&	0.22	&	0.14	&	0.13	&	0.02	&	0.01\\
\textsc{grammar-only}	&	0.01	&	0.03	&	0.14	&	0.12	&	0.18	&	0.17	&	0.01	&	0.01\\
\textsc{morphs-only}	&	0.06	&	0.08	&	0.12	&	0.13	&	0.15	&	0.18	&	0.03	&	0.05\\
\textsc{corpus-grammar}	&	0.01	&	0.01	&	0.14	&	0.17	&	0.12	&	0.08	&	0.01	&	0.01\\
\textsc{corpus-morphs}	&	0.05	&	0.08	&	0.19	&	0.18	&	0.17	&	0.17	&	0.02	&	0.04\\
\textsc{grammar-morphs}	&	0.03	&	0.04	&	0.11	&	0.10	&	0.15	&	0.16	&	0.02	&	0.01\\
\textsc{corpus-grammar-morphs}	&	0.04	&	0.04	&	0.16	&	0.16	&	0.17	&	0.20	&	0.00	&	0.01
\end{tabular}
\caption{BLEU scores for all LLMs and prompt types.}
\label{tab:bleu}
\end{table*}

\begin{table*}[h!]
    \centering
    \textbf{GPT-3.5 Turbo} 
    
    \begin{tabular}{lcccc}
    \\
    \hline
    &	\textbf{None}	&	\textbf{Low}	&	\textbf{Med}	&	\textbf{High}	\\
    \hline
\textsc{baseline}	&	31	&	17	&	2	&	0	\\
\textsc{corpus-only}	&	18	&	23	&	8	&	1	\\
\textsc{grammar-only}	&	20	&	27	&	2	&	1	\\
\textsc{morphs-only}	&	3	&	22	&	16	&	9	\\
\textsc{corpus-grammar}	&	18	&	23	&	9	&	0	\\
\textsc{corpus-morph}	&	2	&	28	&	12	&	8	\\
\textsc{grammar-morph}	&	3	&	29	&	13	&	5	\\
\textsc{corpus-grammar-morph}	&	2	&	27	&	12	&	9	\\
\end{tabular}
\caption{Human quality annotation of GPT-3.5 outputs with automated retrieval (raw counts out of 50) by prompt type.}
\label{tab:fullqualitygpt3}
\end{table*}

\begin{table*}[h!]
    \centering
    \textbf{GPT-4o} 
    
    \begin{tabular}{lcccc}
    \\
    \hline
    &	\textbf{None}	&	\textbf{Low}	&	\textbf{Med}	&	\textbf{High}	\\
    \hline
\textsc{baseline}	&	0	&	10	&	20	&	20\\
\textsc{corpus-only}	&	1	&	16	&	13	&	20\\
\textsc{grammar-only}	&	0	&	17	&	16	&	17\\
\textsc{morphs-only}	&	0	&	13	&	18	&	19\\
\textsc{corpus-grammar}	&	0	&	14	&	17	&	19\\
\textsc{corpus-morph}	&	0	&	10	&	17	&	23\\
\textsc{grammar-morph}	&	0	&	19	&	14	&	17\\
\textsc{corpus-grammar-morph}	&	0	&	9	&	20	&	21\\
\end{tabular}
\caption{Human quality annotation of GPT-4o outputs with automated retrieval (raw counts out of 50) by prompt type.}
\label{tab:fullqualitygpt4}
\end{table*}

\newpage

\begin{table*}
\section{Full Error Counts}
\label{app:errorcounts}

This section contains the full counts of annotated errors by category and prompt type.\\\\
    \centering
  \\
    \textbf{GPT-3.5 Turbo} 
    
    \small
    \renewcommand{\arraystretch}{2}
    \rowcolors{2}{green!80!yellow!50}{green!70!yellow!40}
    \begin{tabular}{p{3cm}cccccccc|c}
    & \textsc{base} & \textsc{c} & \textsc{g} & \textsc{m} & \textsc{c+g} & \textsc{c+m} & \textsc{g+m} & \textsc{c+g+m} & \textsc{Total}\\
\hline
None             & 0 & 1 & 1 & 6 & 0 & 8 & 3 & 5 & 24 \\
Addition         & 0 & 5 & 3 & 14 & 1 & 9 & 10 & 11 & 53 \\
Omission         & 3 & 9 & 2 & 13 & 2 & 5 & 9 & 9 & 52 \\
Substitution - Subject & 0 & 3 & 0 & 7 & 0 & 9 & 9 & 12 & 40 \\
Substitution - TAM      & 0 & 11 & 3 & 17 & 6 & 19 & 19 & 19 & 94 \\
Substitution - Other    & 4 & 9 & 4 & 13 & 6 & 16 & 14 & 13 & 79 \\
Overtranslation         & 1 & 1 & 1 & 4 & 0 & 2 & 3 & 2 & 14 \\
Undertranslation        & 0 & 0 & 0 & 2 & 1 & 2 & 2 & 2 & 9 \\
Target Error - Grammar  & 0 & 1 & 1 & 4 & 2 & 3 & 3 & 1 & 15 \\
Target Error - Coherence        & 0 & 0 & 3 & 5 & 2 & 3 & 7 & 7 & 27 \\
Target Error - Style/Register   & 0 & 3 & 0 & 5 & 2 & 3 & 1 & 2 & 16 \\
Complete Mistranslation         & 30 & 19 & 21 & 2 & 18 & 2 & 2 & 2 & 96 \\
Mistranslation - Lexical Correspondence  & 15 & 13 & 23 & 9 & 21 & 11 & 15 & 10 & 117 \\
Refusal & 1 & 0 & 0 & 0 & 0 & 0 & 0 & 0 & 1 \\
\hline
Total   & 54 & 75 & 62 & 101 & 61 & 92 & 97 & 95 & 637 \\
\end{tabular}
\caption{Human error type annotation of GPT-3.5 outputs with automated retrieval (raw counts, up to 3 errors per sentence) by prompt type.}
\label{tab:fullerrorsgpt35}
\end{table*}

\begin{table*}
    \centering


    \textbf{GPT-4o} 
    
    \small
    \renewcommand{\arraystretch}{2}
    \rowcolors{2}{green!80!yellow!50}{green!70!yellow!40}
    \begin{tabular}{p{3cm}cccccccc|c}
    & \textsc{base} & \textsc{c} & \textsc{g} & \textsc{m} & \textsc{c+g} & \textsc{c+m} & \textsc{g+m} & \textsc{c+g+m} & \textsc{Total}\\
\hline
None                         & 15 & 16 & 10 & 16 & 13 & 19 & 14 & 18 & 121 \\
Addition                     & 2  & 5  & 7  & 5  & 4  & 1  & 6  & 4  & 34 \\
Omission                     & 8  & 7  & 6  & 7  & 6  & 3  & 5  & 5  & 47 \\
Substitution - Subject       & 1  & 2  & 0  & 1  & 2  & 1  & 2  & 2  & 11 \\
Substitution - Other         & 22 & 24 & 22 & 18 & 19 & 18 & 17 & 20 & 160 \\
Substitution - TAM           & 16 & 17 & 19 & 12 & 13 & 10 & 11 & 9  & 107 \\
Overtranslation              & 2  & 1  & 0  & 2  & 2  & 2  & 1  & 2  & 12 \\
Undertranslation             & 6  & 1  & 3  & 1  & 3  & 0  & 1  & 2  & 17 \\
Target Error - Grammar       & 1  & 3  & 4  & 4  & 1  & 2  & 6  & 1  & 22 \\
Target Error - Coherence     & 1  & 3  & 4  & 5  & 4  & 5  & 9  & 5  & 36 \\
Target Error - Style/Register & 1  & 2  & 3  & 4  & 4  & 2  & 4  & 3  & 23 \\
Complete Mistranslation      & 0  & 1  & 0  & 0  & 0  & 0  & 0  & 0  & 1 \\
Mistranslation - Lexical Correspondence & 4  & 3  & 5  & 6  & 6  & 6  & 9  & 4  & 43 \\
\hline
Total                         & 79 & 85 & 83 & 81 & 77 & 69 & 85 & 75 & 634 \\
\end{tabular}
\caption{Human error type annotation of GPT-4o outputs with automated retrieval (raw counts, up to 3 errors per sentence) by prompt type.}
\label{tab:fullerrorsgpt4}
\end{table*}



\end{document}